\documentclass[a4paper]{article}
\usepackage{idsiatechrep}      
\usepackage{graphicx}
\usepackage{natbib}

\title{High-Performance Neural Networks\\ for Visual Object Classification}
\author{Dan C. Cire\c{s}an, Ueli Meier, Jonathan Masci,\\ Luca M. Gambardella and J{\"u}rgen Schmidhuber}
\date{January 2011}

\reportnumber{01-11}

\reportaddnote{This work was partially supported by the Swiss Commission for Technology and Innovation (CTI), Project n. 9688.1 IFF: Intelligent Fill in Form.}

\begin{document}
\makecover         
\maketitle

\begin{abstract}
We present a fast, fully parameterizable GPU implementation of
Convolutional Neural Network variants. Our feature extractors are
neither carefully designed nor pre-wired, but rather learned in a
supervised way. Our deep hierarchical architectures achieve the
best published results on benchmarks for object classification (NORB, CIFAR10)
and handwritten digit recognition (MNIST),
with error rates of  2.53\%, 19.51\%, 0.35\%, respectively. 
Deep nets trained by simple
back-propagation perform better than more shallow ones. Learning is
surprisingly rapid. NORB is completely trained within five
epochs. Test error rates on MNIST drop to 2.42\%, 0.97\% and 0.48\%
after 1, 3 and 17 epochs, respectively. 

\end{abstract}


\section{Introduction}
The human visual system efficiently recognizes and localizes objects
within cluttered scenes. For artificial systems, however, this is
still difficult, due to viewpoint-dependent object variability,
and the high in-class variability of many object types.  Deep
hierarchical neural models roughly mimick the nature of mammalian
visual cortex, and by community consensus are among the most promising
architectures for such tasks. The most successful hierarchical object
recognition systems all extract localized features from input
images, convolving image patches with filters. Filter responses are
then repeatedly sub-sampled and re-filtered, resulting in a deep
feed-forward network architecture whose output feature vectors are
eventually classified. One of the first hierarchical neural systems was the Neocognitron
\citep{fukushima:1980}  which inspired many of the more recent
variants.

Unsupervised learning methods applied to patches of natural images
tend to produce localized filters that resemble off-center-on-surround
filters, orientation-sensitive bar detectors, Gabor filters
\citep{schmidhuber:1996,olshausen:1997,hoyer:2000}. These findings
in conjunction with experimental studies of the visual cortex justify the
use of such filters in the so-called \textit{standard model} for
object recognition \citep{riesenhuber:1999,serre:2005,mutch:2008},
whose filters are fixed, in contrast to those of Convolutional Neural
Networks (CNNs) \citep{lecun:1998,behnke:2003,simard:2003}, whose
weights (filters) are randomly initialized and changed in a supervised
way using back-propagation (BP).

Despite the hardware progress of the past decades, computational
speed is still a limiting factor for CNN architectures characterized
by many building blocks typically set by trial and error.  To
systematically test the impact of various architectures on
classification performance, we present a fast CNN implementation on
Graphics Processing Units (GPUs). Previous GPU implementations of CNNs
\citep{chellapilla:2006b,uetz:2009} were hard-coded to satisfy GPU
hardware constraints, whereas our implementation is flexible and fully
online (i.e., weight updates after each image). It allows for training
large CNNs within days instead of months, such that we can investigate the
influence of various structural parameters by exploring large
parameter spaces \citep{pinto:2009} and performing error
analysis on repeated experiments.

We evaluate various networks on the handwritten digit benchmark MNIST \citep{lecun:1998}
 and two image classification benchmarks: NORB \citep{lecun:2004} and CIFAR10 \citep{krizhevsky:2009}.

\section{Convolutional neural networks}

CNNs are hierarchical neural networks whose convolutional layers
alternate with subsampling layers, reminiscent of simple and complex
cells in the primary visual cortex \citep{wiesel:1959}. 
CNNs vary in how convolutional and subsampling layers are realized
and how the nets are trained. The CNN architecture considered in
this study differs from the one of \citet{simard:2003} in the sense that
after each CNN-layer an optional max-pooling
layer \citep{scherer:2010} can be used. Here we give a complete description
of this independent implementation
(Fig. \ref{Fig:CNN-architecture}).

\subsection{Image processing layer}
The image processing layer is an optional pre-processing layer of
predefined filters that are kept fixed during training. Thus 
additional information besides the raw input image can be provided to
the network, such as edges and gradients. In particular, we find that
a contrast-extracting layer \citep{fukushima:2003} helps to improve
the recognition rate for NORB.

\subsection{Convolutional layer}
\begin{figure}[ht]
\hfill
\begin{center}
\includegraphics[width=\columnwidth]{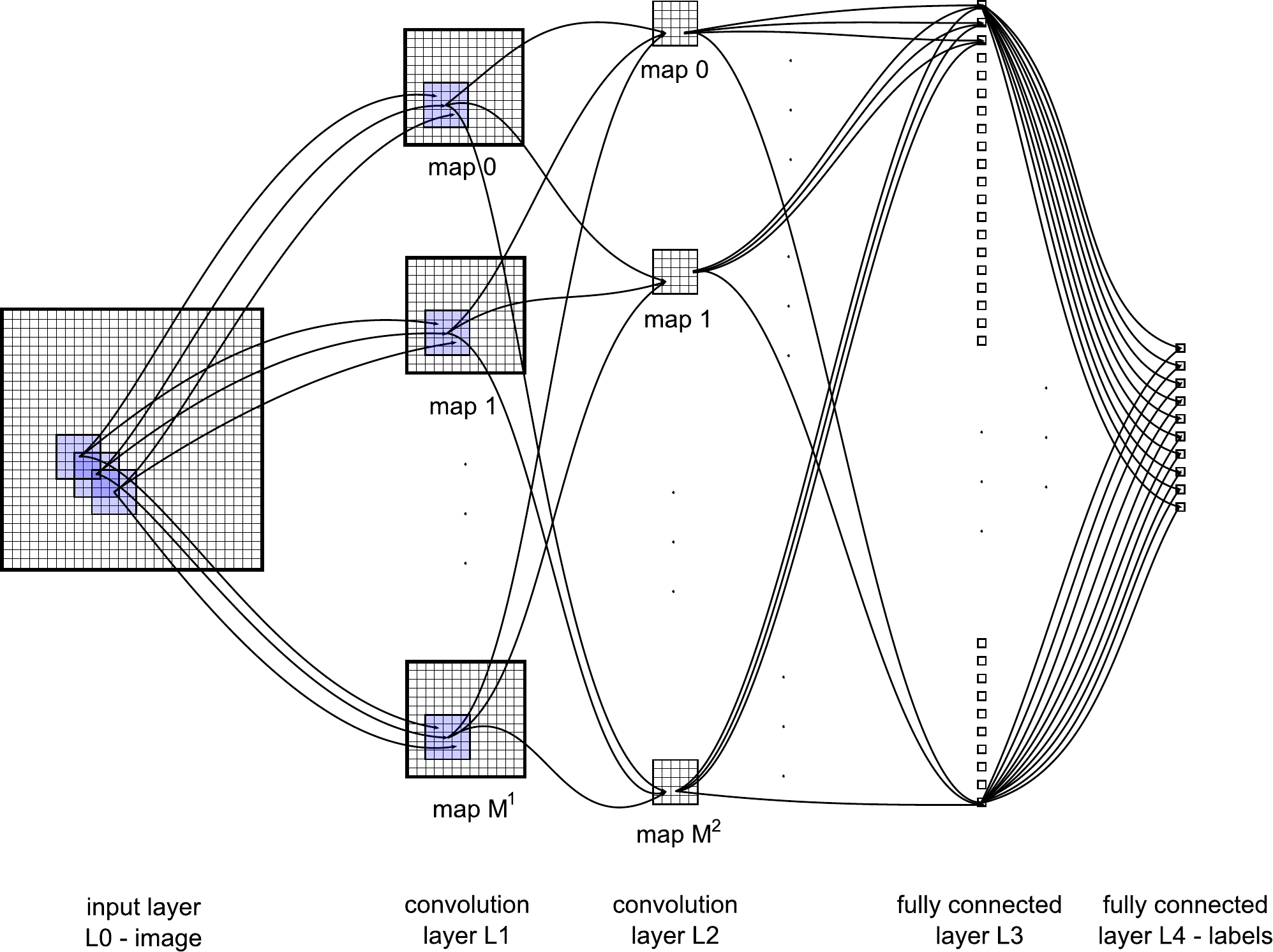}
\end{center}
\caption{Architecture of a convolutional neural network.  In this
case, the convolutional layers are fully connected. Both convolutional
layers use a kernel of 5 x 5 and skipping factors of 1.}
\label{Fig:CNN-architecture}
\end{figure}

A convolutional layer is parametrized by the size and the number of the
maps, kernel sizes, skipping factors, and the connection table. Each layer
has $M$ maps of equal size ($M_x$, $M_y$). A kernel (blue rectangle in
Fig~\ref{Fig:CNN-architecture}) of size ($K_x$, $K_y$) is shifted
over the valid region of the input image (i.e. the kernel has to be
completely inside the image). The skipping factors $S_x$ and $S_y$
define how many pixels the filter/kernel skips in x- and y-direction
between subsequent convolutions. The size of the output map is then
defined as:
\begin{equation} 
M^n_x = \frac{M^{n-1}_x-K_x^n}{S_x^n+1}+1; \quad M^n_y = \frac{M^{n-1}_y-K_y^n}{S_y^n+1}+1\\
\end{equation}
\noindent where index $n$ indicates the layer. Each map in layer
$L^n$ is connected to at most $M^{n-1}$ maps in layer
$L^{n-1}$. Neurons of a given  map share their weights but have different
receptive fields.

\subsection{Max-pooling layer} 
The biggest architectural difference between our implementation and
the CNN of \citet{lecun:1998} is the use of a max-pooling layer
instead of a sub-sampling layer. No such layer is used by
\citet{simard:2003} who simply skips nearby pixels prior to
convolution, instead of pooling or averaging. \citet{scherer:2010}
found that max-pooling can lead to faster convergence, select superior
invariant features, and improve generalization. The output of the
max-pooling layer is given by the maximum activation over
non-overlapping rectangular regions of size ($K_x$,
$K_y$). Max-pooling enables position invariance over larger local
regions and downsamples the input image by a factor of $K_x$ and $K_y$
along each direction.

\subsection{Classification layer} 
Kernel sizes of convolutional filters and max-pooling rectangles as
well as skipping factors are chosen such that either the output maps
of the last convolutional layer are downsampled to 1 pixel per map, or
a fully connected layer combines the outputs of the topmost convolutional
layer into a 1D feature vector. The top layer is always fully
connected, with one output unit per class label.
 
\section{GPU implementation}

The latest generation of NVIDIA GPUs, the 400 and 500 series (we use
GTX 480 \& GTX 580), has many advantages over older GPUs, most notably
the presence of a R/W L2 global cache for device memory. This
permits faster programs and
simplifies writing the code. In fact, the corresponding transfer of
complexity into hardware alleviates many software and optimization
problems. Our experiments show that the CNN program becomes 2-3 times
faster just by switching from GTX 285 to GTX 480. 

Manual optimization
of CUDA code is very time-consuming and error prone. We optimize for
the new architecture, relying on the L2 cache for many of the device
memory accesses, instead of manually writing code that uses textures
and shared memory. Code obtained by this pragmatic strategy is fast
enough.  We use
the following types of optimization: pre-computed expressions,
unrolled loops within template kernels, strided matrices to obtain
coalesced memory accesses and registers wherever possible.
Additional manual optimizations are possible in case future image
classification problems will require even more computing power.

\subsection{Data structures}

Both outputs $y$ and deltas $\delta$ of layer $L^n$ are 2D
strided. Their original size is $M_x\times MM_y$, but they are
horizontally strided with a pitch of 32 floats (we use this stride for
all 2D data), resulting in coalesced memory accesses. The vertical
stride avoids additional bounding tests in CUDA kernels.

All connections between maps of consecutive layers $L^{n-1}$ and $L^n$
are stored in matrix $C^n$. Each row of $C^n$ contains all connections
that feed into a particular map in layer $L^n$. Because we aim for a
flexible architecture with partially connected layers, in the first
column we store the number of previous connections. This index is
useful for Forward Propagation (FP) and Adjusting Weights (AW) CUDA
kernels. The second column stores the number of connections, followed
by corresponding indices of maps in $L^{n-1}$ connected to
the current map.

For BP and FP, analogous information about connections is needed.
We therefore store backward connections in $C_{BP}$. AW requires a
list of all map connections (see Subsection~\ref{Sub:WA}), stored as
an array of map index pairs. Dealing with biases in BP kernel requires
to know where the weights of particular connections start; this information is
stored in a 2D array $WIDX_{BP}$ of size $M^n \times M^{n-1}$.

\subsection{Forward propagation}
A straightforward way of parallelizing FP is to assign a thread block
to each map that has to be computed. For maps bigger than 1024
neurons, the job is further split into smaller blocks by assigning a
block to each line of the map, because the number of
threads per block is limited (1024 for GTX 480). A one to one
correspondence between threads and the map's neurons is
assumed. Because of weight sharing, threads inside a block can access
data in parallel, in particular the same weights and inputs from the
previous layer. Each thread starts by initializing its sum with the
bias, then loops over all map connections, convolving the
appropriate patch of the input map with the corresponding kernel. The
output is obtained by passing the sum through a scaled tanh activation
function, and then written to device memory.

\subsection{Backward propagation}
BP of deltas can be done in two ways: by pushing or by
pulling. Pushing deltas means taking each delta from the current layer
and computing the corresponding deltas for the previous layer. For an
architecture with shared weights this has the disadvantage of being
hard to code. Each delta from the current layer contributes to
many deltas in the previous layer, which translates into a lot of
programming. There are two ways of avoiding this: either writing
partial deltas to a separated block of memory and then putting
everything together by calling another kernel (slow because of a
tremendous increase in the number of memory accesses, and the need of another
kernel), or using atomic writes  (to avoid data hazards) to update
deltas (very slow because many writings are serialized). 
We implement pulling
deltas, which has almost none of the above speed-limiting drawbacks, but
is a bit more complicated. 

\begin{figure}[ht]
\hfill
\begin{center}
\includegraphics[width=0.75\columnwidth]{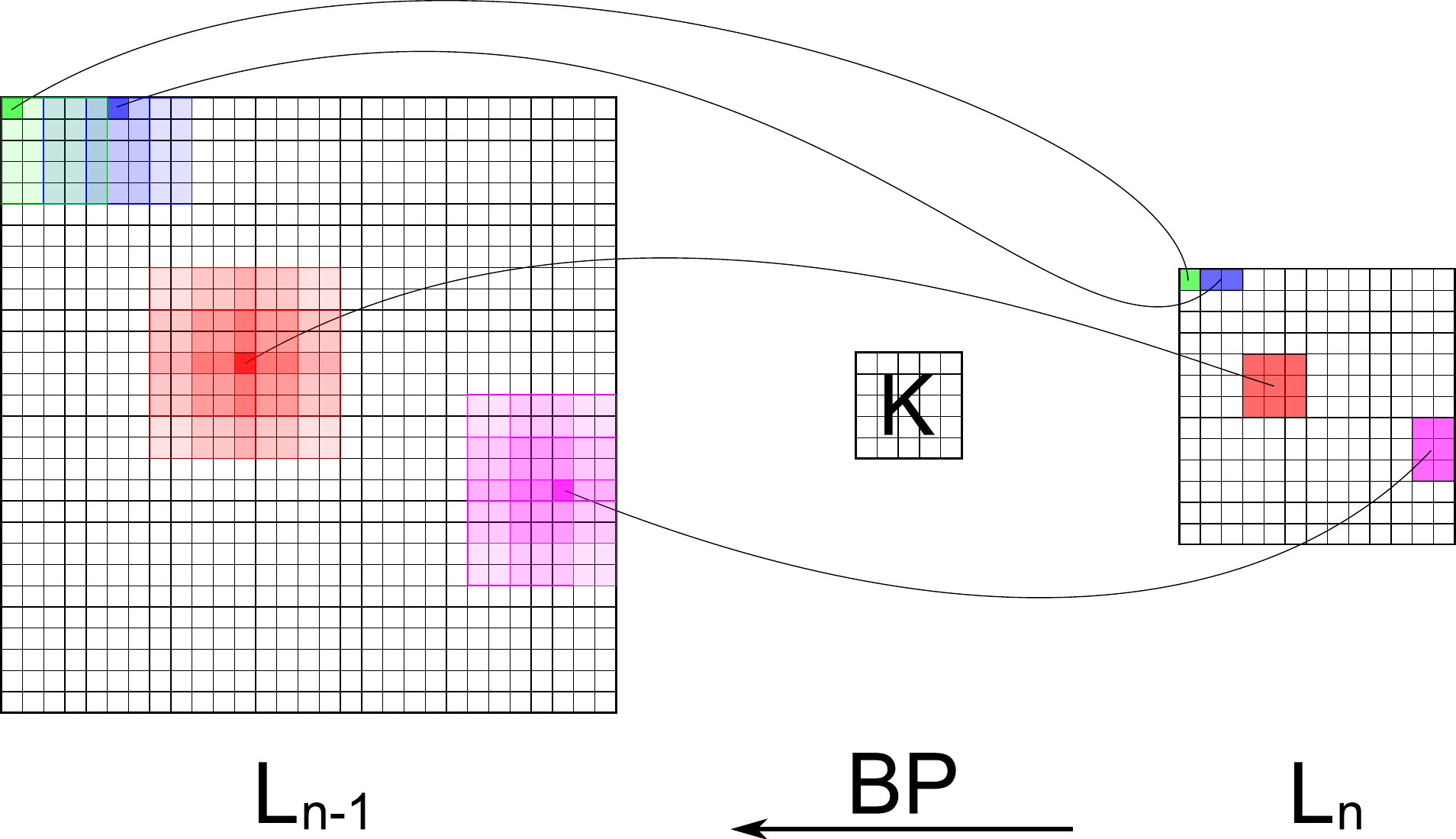}
\end{center}
\caption{Back propagating deltas. A connection between two maps from
two consecutive layers is displayed. The map in $L^{n-1}$ has 29 x 29
neurons; the map in $L^n$ has 13 x 13 neurons. They are linked through
a 5 x 5 kernel $K$. Skipping factors of $S_x=1$ and $S_y=1$ are
assumed. Arrows and colors depict the correspondence between neurons
in $L^{n-1}$ and their sources in $L^n$.}
\label{Fig:BPdeltas}
\end{figure}

The (uni- or bi-dimensional) thread grid assigns a (bi- or
uni-dimensional) thread block to each map in the previous layer and a
thread to each neuron in every map. Similar to FP, for maps with more
than 1024 neurons, the 2D grid is further split into smaller 1D blocks
by assigning a 2D block to each row of the map. Each thread computes
the delta of its corresponding neuron by pulling deltas from the
current layer. For every neuron in the previous layer we have to
determine the list of neurons in the current layer which are connected
to it. Let us consider neuron $(i,j)$ from a map in layer
$L^{n-1}$, and then assume that $(x,y)$ are the coordinates of neurons
in maps of $L^n$ that contribute to the delta of neuron
$(i,j)$. The $(x,y)$ neuron is connected to kernel size number 
neurons ($K_x \times K_y$) from each connected map in the previous
layer. The indices in $L^{n-1}$ of the neurons connected through a
kernel to the $(x,y)$ neuron are:
\begin{eqnarray}
  x(S_x+1)  &\le  i \le& x(S_x+1)+K_x-1 \nonumber,\\
  y(S_y+1)  &\le  j \le& y(S_y+1)+K_y-1 \nonumber.
\end{eqnarray}
We can now compute the inequalities for $(x,y)$:
\begin{eqnarray}
   \frac{i-K_x+1}{S_x+1} &\le   x   \le&  \frac{i}{S_x+1} \nonumber, \\
   \frac{j-K_y+1}{S_y+1} &\le   y   \le&  \frac{j}{S_y+1} \nonumber.
\end{eqnarray}
Because $(x,y)$ has to be inside the map, the final inequalities are:
\begin{eqnarray}
  \max{\left ( \left \lceil \frac{i-K_x+1}{S_x+1} \right \rceil,0 \right )} &\le x \le & \min{\left ( \left \lfloor \frac{i}{S_x+1} \right \rfloor,M_x-1 \right )}\nonumber, \\
 \max{\left ( \left \lceil \frac{j-K_y+1}{S_y+1} \right \rceil,0 \right )} &\le y \le &\min{\left ( \left \lfloor \frac{j}{S_y+1} \right \rfloor,M_y-1 \right )}.\nonumber
\end{eqnarray}
The above inequalities state that the delta of neuron $(i,j)$ from
$L^{n-1}$ is computed from deltas of neurons in a rectangular area in
maps of $L^n$ (Fig.~\ref{Fig:BPdeltas}). After summing up the deltas,
each thread multiplies the result by the derivative of the activation
function.

\subsection{Adjusting weights}
FP and BP have a grid on the list of maps, but the AW
thread grid is on the list of kernels (filters) between maps of
two consecutive layers. The 1D grid has a block for each
connection between two maps. Thread blocks are 2D, with a
corresponding thread for every kernel weight. The bias weight
is included as an entire row of threads, thus requiring thread blocks
to have $(K_x+1) \times K_y$ threads. Most of the time these additional
$K_y$ threads will do nothing, thread (0,0) being activated only for
blocks that have to process the bias.
\label{Sub:WA}

\section{Experiments}

We use a system with a Core i7-920 (2.66GHz), 12 GB DDR3 and four
graphics cards: 2 x GTX 480 and 2 x GTX 580.  The correctness of the
CPU version is checked by comparing the analytical gradient with its
finite difference approximation. On GPU this is not
possible because all computations are performed with single precision
floating point numbers. Hence the GPU implementation's correctness is
checked by comparing its results to those of a randomly initialized net after
training it for several epochs on the more accurate 
CPU version. Obtaining identical results 
after trillions of operations is
a strong indication of correctness.

The implemented CNN's plain feed-forward architecture is 
trained using on-line gradient descent. All images from the training
set are used for training and also for validation. If deformations are
enabled, only the images from the training set will be deformed. Weights
are initialized according to a uniform random distribution in the range
$[-0.05,0.05]$. Each neuron's activation function is a scaled
hyperbolic tangent: $y(a)=1.7159\tanh(0.6666a)$ \citep{lecun:1998}.

We pick the trained CNN with the lowest validation error, and evaluate
it on the test set (Test for best Validation - TfbV). 
The best test error (bT) is also listed for all experiments. The reported
computation times per epoch include training, validation and testing
as well as all data transfers.

\subsection{Experiments on MNIST}

For the MNIST dataset the networks are trained on deformed images,
continually generated in on-line fashion. Affine (translation,
rotation, scaling, horizontal shearing) and elastic deformations
\citep{simard:2003} are combined.  We use a variable learning rate
that shrinks by a multiplicative constant after each epoch, from
$10^{-3}$ down to $3\cdot10^{-5}$ after 500 epochs.

\begin{table}[!t]
\caption{Error rates on MNIST test set for randomly connected CNNs
with 2 to 6 convolutional layers with M Maps and an optional fully
connected layer with N neurons. Various kernel sizes and skipping
factors were used.}
\label{table_MNIST_random}
\centering
\begin{tabular}{c|cc}
\#M, \#N 	&		bT	&	TfbV\\
 in Hidden Layers&[\%]&[\%]\\
\hline
20M-60M & 0.95 & 1.02\\
20M-60M-150N & 0.50 & 0.55\\
20M-60M-100M-150N & 0.33  & 0.38\\
20M-40M-60M-80M-100M-120M-150N & \bf{0.27} & \bf{0.35}\\
\end{tabular}
\end{table} 

Fully connected convolutional layers lead to an exploding number of
network connections and weights, making training of big and deep CNNs
for hundreds of epochs impractical even on GPUs. Partial connectivity
alleviates this problem and is also biologically more plausible.
We reduce the number of connections between convolutional
layers in a random way. Table~\ref{table_MNIST_random} lists results
of various networks with 2 to 7 hidden layers with random
connections. Additional layers result in better networks, the best one
achieving a test error of 0.35\% for best validation and a best test
error of 0.27\%. The best previous CNN result on MNIST is 0.40\%
\citep{simard:2003}. A 0.35\% error rate was recently also obtained by a big,
deep MLP \citep{ciresan:2010} with many more free parameters.
Deeper nets require more computation time to complete an epoch, but we
observe that they also need fewer epochs to achieve good test
errors. The deepest CNN from Table~\ref{table_MNIST_random} reaches
2.42\%, 0.97\% and 0.48\% after one, three and seventeen epochs,
respectively. On the other hand, the network with 4 instead of 7
hidden layers reaches 4.71\%, 1.58\%, 0.68\% after one, three and
seventeen epochs, achieving a test error below 0.50\% after only 34
epochs. This shows once more that deep networks, contrary to common
belief, can be trained successfully by back-propagation. Despite the
numerous free parameters, deep networks seem to learn faster (better
recognition rates after fewer epochs) than shallow ones.

We consider MNIST an almost solved problem. The remaining errors stem from digits that are  ambiguous or  miss parts. 

\subsection{Experiments on NORB} 

NORB contains stereo images of 3D objects. Hence there are two maps on the input layer. Rotation, 
scaling, shearing and elastic distortions seem to have a negative
impact on generalization. These deformations improve recognition rates
for digits that are intrinsically 2D \citep{ciresan:2010}, but seem inadequate for 3D
objects.

Initial experiments on NORB show that unlike with MNIST where we use
deformations, the CNN needs only 3 to 6 epochs to reach zero
validation error. This allows us to quickly run numerous repetitive
experiments with huge networks with hundreds of
maps per layer. We decided to use a CNN with five hidden
layers: layer1, a convolutional layer with 300 maps, kernel size
$6\times6$ and skipping factors $1\times1$; layer2, a max-pooling
layer over a $2\times2$ region; layer3, a convolutional layer with 500
maps, kernel size $4\times4$, skipping factors $0\times0$; layer4, a
max-pooling layer over a $4\times4$ region; layer5, a fully connected
layer with 500 neurons. The learning rate is initialized by 0.001
and multiplied by 0.95 after every epoch.

Table \ref{table_NORB} summarizes the results of four different
experiments by switching on/off translation as well as the fixed image
processing layer. We report the average error rate as well as the
standard deviation of N independent runs with identical architectures but
different weight initializations. For the first experiment without
translation and no image processing (IP), an average test error rate of
7.86\% is obtained. With additional translations of at most 5\%, the
average error rate drops to 4.71\%, contradicting the common
belief that CNNs are translation invariant. These results are on par
or better than others in the literature: 5.90\%
error rate for a combination of CNNs and SVMs \citep{lecun:2004}
and 5.20\% error rate for restricted Boltzman machines
\citep{nair:2009}.

The best previously published result on NORB (2.87\%) was obtained by a
hierarchical neural network which to every convolutional layer 
provides a subsampled version plus edge information of the
original image  \citep{uetz:2009}. This
motivated us to implement a pre-processing layer with fixed
filters. We tried simple edge masks (Sobel, Scharr)  but obtained best results
with a  contrast-extraction layer \citep{fukushima:2003}  realized by Mexican
hat-shaped filters of size $21\times21$, 
one with a concentric on-center receptive field and one with
a concentric off-center receptive field, similar to the filters automatically created  by 
unsupervised {\em Predictability Minimization} \citep{schmidhuber:1992} applied to natural images \citep{schmidhuber:1996}.
The first filter extracts
positive contrast in brightness, whereas the latter extracts negative
contrast. Each image from the original NORB is filtered, consequently
the input of the CNN has six maps: the original image plus
the positive and negative contrast for each of the two stereo
channels. Using such a pre-processing layer results in lower average
error rates, 3.94\% without translation and 2.53\% with
translation. This result improves the previous state of the art on NORB
\citep{uetz:2009}.

\begin{table}[!t]
\caption{Average error rates and standard deviations of N runs for a five hidden layer CNN on the NORB test set (see text for details).}
\label{table_NORB}
\centering
\begin{tabular}{ccccc}
trans. [\%]	&	IP 	&	TfbV [\%]& runs	& time/epoch [s]		\\
\hline
0		&	no	&	7.86 $\pm$	0.55	& 50	&	1141		\\
5		&	no	&	4.71 $\pm$	0.57	& 50	&	1563		\\
0		&	yes	&	3.94 $\pm$  0.48	& 50	&	1658		\\
5		&	yes	&	{\bf 2.53} $\pm$  0.40	& 100	&	2080		\\
\end{tabular}
\end{table}

Experience with other image datasets tells us that NORB is unusual. 
The training set has only five instances per
class. The resulting  poor training set variability makes the nets learn quickly but 
generalize badly. NORB is
the only dataset that profits  from a fixed
pre-processing layer in a substantial way. For MNIST and CIFAR10 such 
pre-processing has little or no effect. It is also worth noting that NORB's standard
error rate deviation is bigger than CIFAR10's
(see Tables \ref{table_NORB} and \ref{table_CIFAR10}). Identical nets
with different initializations do not produce very consistent results. The
best net had an error rate of 1.72\%, the worst 3.69\%.

\subsection{Experiments on CIFAR 10}

CIFAR10 is a collection of  natural color images of 32x32 pixels. It
contains 10 classes, each of them with 5000 samples in the training
set and 1000 in the test set. The images greatly vary inside
each class. They are not necessarily centered, may contain only
parts of the object, and have varying backgrounds. All of this
makes CIFAR10 the hardest problem addressed in this paper. The
CNN has three maps, one for each color channel
(RGB). The CIFAR10 images are relatively small in comparison to NORB's, and
force us to use small kernels. The tested CNNs differ only in
the number of maps per convolutional and max-pooling layer. All have
eight hidden layers: layer1, a convolutional layer with $3\times3$
kernels and skipping factor of $0$; layer2, a max-pooling layer over
a $3\times3$ region; layer3, a convolutional layer with $3\times3$
kernels and skipping factors of $0\times0$; layer4, a max-pooling over
a $2\times2$ region; layer5, a convolutional layer with $3\times3$
kernels and a skipping factors of $0\times0$; layer6, a max pooling
layer over a $2\times2$ region; layer7, a fully connected layer with
$300$ neurons; layer8, a fully connected layer with $100$ neurons.

Like for MNIST, the learning rate is  initialized by
0.001 and multiplied by 0.993 after every epoch. Results in
Table \ref{table_CIFAR10} show that without translation the error rate
does not drop below 28\%; adding edge information does not help at
all. Translations have a very positive effect, decreasing the error
rate to almost 20\%. Contrast extraction filters are better than
the Sobel/Scharr filters but still worse than no pre-processing
layer at all. Despite some CNN-inherent translation invariance, additional
training image translations cause better generalization;
additional image processing proved useless though.

\begin{table}[!t]
\caption{Average error rates and standard deviations for N runs of an
eight hidden layer CNN on the CIFAR10 test set (see text for
details). The first five nets have 100 maps per convolutional and
max-pooling layer, whereas the sixth, seventh and eighth have 200, 300
and 400 maps per hidden layer, respectively. IP - image processing layer: edge - $3\times3$ Sobel
and Scharr filters; hat - $13\times13$ positive and negative contrast
extraction filters.}
\label{table_CIFAR10}
\centering
\begin{tabular}{cccccc}
trans. [\%]	&	maps		&	IP 	&	TfbV [\%]& runs	& time/epoch [s]		\\
\hline
0			&	100		&	no	  &	28.87 $\pm$ 0.37		& 11		&	93		\\
0			&	100		&	edge	&	29.11 $\pm$ 0.36		& 15		&	104		\\
5			&	100		&	no	  &	20.26 $\pm$ 0.21		& 11		&	111		\\
5			&	100		&	edge	&	21.87 $\pm$ 0.57		& 5		&	120		\\
5			&	100		&	hat	  &	21.44 $\pm$ 0.44		& 4		&	136		\\
\hline
5			&	200		&	no	&	19.90 $\pm$ 0.16		& 5		&	248		\\
5			&	300		&	no	&	{\bf 19.51} $\pm$ 0.18	& 5		&	532		\\
5			&	400		&	no	&	19.54 $\pm$ 0.16		& 5		&	875		\\
\end{tabular}
\end{table}

To see if bigger nets are better, we increase the number of maps per
layer from 100 to 200, 300 and 400, respectively (last three rows in
Tab.~\ref{table_CIFAR10}). Training time increases exponentially, but
the test error decreases, reaching a minimum for nets with 300 maps
per layer. Our 19.51\% error rate is better than the previous state of
the art for this dataset, 20.40\% \citep{lee:2010} and 25.50\%
\citep{yu:2010}. Unlike \citet{lee:2010}, however, we use the original
images without any particular input normalization. Note that the error
rate standard deviations are smaller than those obtained on NORB, that is,
different initializations yield consistent results.

\subsection{Speedup factor of GPU code}
The GPU code scales well with network size. For small nets the
speedup is small (but still over 10) since they fit better inside the
CPU cache, and GPU resources are underutilized. For huge nets (ex:
Table~\ref{table_NORB}) the GPU implementation is more than 60 times
faster than a compiler-optimized CPU version. Given the flexibility of
our GPU version, this is a significant speedup. One epoch takes 35 GPU
minutes but more than 35 CPU hours.

\section{Conclusion}
We presented high-performance GPU-based CNN variants trained by
on-line gradient descent, with sparse random connectivity,
computationally more efficient and biologically more plausible than
fully connected CNNs. Principal advantages include state-of-the-art
generalization capabilities, great flexibility and speed. All
structural CNN parameters such as input image size, number of hidden
layers, number of maps per layer, kernel sizes, skipping factors and
connection tables are adaptable to any particular application.  We
applied our networks to benchmark datasets for digit recognition
(MNIST), 3D object recognition (NORB), and natural images
(CIFAR10). On MNIST the best network achieved a recognition test error rate
of 0.35\%, on NORB 2.53\% and on CIFAR10 19.51\%. Our results are
raising the bars for all three benchmarks.  Currently the particular
CNN types discussed in this paper seem to be the best adaptive image
recognizers, provided there is a labeled dataset of sufficient
size. No unsupervised pretraining is required. Good results require
big and deep but sparsely connected CNNs, computationally prohibitive
on CPUs, but feasible on current GPUs, where our implementation is 10
to 60 times faster than a compiler-optimized CPU version.


\section*{Acknowledgment}

This work was partially funded by the Swiss Commission for Technology and Innovation (CTI),
Project n. 9688.1 IFF: Intelligent Fill in Form.


\begin{thebibliography}{}

\bibitem[\protect\citeauthoryear{Behnke}{2003}]{behnke:2003}
S.~Behnke.
\newblock {\em Hierarchical Neural Networks for Image Interpretation}, volume
  2766 of {\em Lecture Notes in Computer Science}.
\newblock Springer, 2003.

\bibitem[\protect\citeauthoryear{Chellapilla \bgroup \em et al.\egroup
  }{2006}]{chellapilla:2006b}
K.~Chellapilla, S.~Puri, and P.~Simard.
\newblock High performance convolutional neural networks for document
  processing.
\newblock In {\em International Workshop on Frontiers in Handwriting
  Recognition}, 2006.

\bibitem[\protect\citeauthoryear{Cire\c{s}an \bgroup \em et al.\egroup
  }{2010}]{ciresan:2010}
D.~C. Cire\c{s}an, U.~Meier, L.~M. Gambardella, and J.~Schmidhuber.
\newblock Deep big simple neural nets for handwritten digit recogntion.
\newblock {\em Neural Computation}, 22(12):3207--3220, 2010.

\bibitem[\protect\citeauthoryear{Coates \bgroup \em et al.\egroup
  }{2010}]{lee:2010}
A.~Coates, H.~Lee, and A.~Ng.
\newblock An analysis of single-layer networks in unsupervised feature
  learning.
\newblock In {\em Advances in Neural Information Processing Systems}, 2010.

\bibitem[\protect\citeauthoryear{Fukushima}{1980}]{fukushima:1980}
K.~Fukushima.
\newblock Neocognitron: A self-organizing neural network for a mechanism of
  pattern recognition unaffected by shift in position.
\newblock {\em Biological Cybernetics}, 36(4):193--202, 1980.

\bibitem[\protect\citeauthoryear{Fukushima}{2003}]{fukushima:2003}
K.~Fukushima.
\newblock {Neocognitron for handwritten digit recognition}.
\newblock {\em Neurocomputing}, 51:161--180, 2003.

\bibitem[\protect\citeauthoryear{Hoyer and Hyv{\"a}rinen}{2000}]{hoyer:2000}
P.~O. Hoyer and A.~Hyv{\"a}rinen.
\newblock Independent component analysis applied to feature extraction from
  colour and stero images.
\newblock {\em Network: Computation in Neural Systems}, 11(3):191--210, 2000.

\bibitem[\protect\citeauthoryear{Krizhevsky}{2009}]{krizhevsky:2009}
A.~Krizhevsky.
\newblock Learning multiple layers of features from tiny images.
\newblock Master's thesis, Computer Science Department, University of Toronto,
  2009.

\bibitem[\protect\citeauthoryear{LeCun \bgroup \em et al.\egroup
  }{1998}]{lecun:1998}
Y.~LeCun, L.~Bottou, Y.~Bengio, and P.~Haffner.
\newblock Gradient-based learning applied to document recognition.
\newblock {\em Proceedings of the IEEE}, 86(11):2278--2324, November 1998.

\bibitem[\protect\citeauthoryear{LeCun \bgroup \em et al.\egroup
  }{2004}]{lecun:2004}
Y.~LeCun, F.-J. Huang, and L.~Bottou.
\newblock Learning methods for generic object recognition with invariance to
  pose and lighting.
\newblock In {\em Proc. of Computer Vision and Pattern Recognition Conference},
  2004.

\bibitem[\protect\citeauthoryear{Mutch and Lowe}{2008}]{mutch:2008}
J.~Mutch and D.~G. Lowe.
\newblock Object class recognition and localization using sparse features with
  limited receptive fields.
\newblock {\em Int. J. Comput. Vision}, 56(6):503--511, 2008.

\bibitem[\protect\citeauthoryear{Nair and Hinton}{2009}]{nair:2009}
V.~Nair and G.~E. Hinton.
\newblock 3d object recognition with deep belief nets.
\newblock In {\em Advances in Neural Information Processing Systems}, 2009.

\bibitem[\protect\citeauthoryear{Olshausen and Field}{1997}]{olshausen:1997}
B.~A. Olshausen and D.~J. Field.
\newblock Sparse coding with an overcomplete basis set: A strategy employed by
  v1?
\newblock {\em Vision Research}, 37(23):3311--3325, December 1997.

\bibitem[\protect\citeauthoryear{Pinto \bgroup \em et al.\egroup
  }{2009}]{pinto:2009}
N.~Pinto, D.~Doukhan, J.~J. DiCarlo, and D.~D. Cox.
\newblock {A high-throughput screening approach to discovering good forms of
  biologically inspired visual representation.}
\newblock {\em PLoS computational biology}, 5(11):e1000579, November 2009.

\bibitem[\protect\citeauthoryear{Riesenhuber and
  Poggio}{1999}]{riesenhuber:1999}
M.~Riesenhuber and T.~Poggio.
\newblock Hierarchical models of object recognition in cortex.
\newblock {\em Nat. Neurosci.}, 2(11):1019--1025, 1999.

\bibitem[\protect\citeauthoryear{Scherer \bgroup \em et al.\egroup
  }{2010}]{scherer:2010}
D.~Scherer, A.~M{\"u}ller, and S.~Behnke.
\newblock Evaluation of pooling operations in convolutional architectures for
  object recognition.
\newblock In {\em International Conference on Artificial Neural Networks},
  2010.

\bibitem[\protect\citeauthoryear{Schmidhuber \bgroup \em et al.\egroup
  }{1996}]{schmidhuber:1996}
J.~Schmidhuber, M.~Eldracher, and B.~Foltin.
\newblock Semilinear predictability minimization produces well-known feature
  detectors.
\newblock {\em Neural Computation}, 8(4):773--786, 1996.

\bibitem[\protect\citeauthoryear{Schmidhuber}{1992}]{schmidhuber:1992}
J.~Schmidhuber.
\newblock Learning factorial codes by predictability minimization.
\newblock {\em Neural Computation}, 4(6):863--879, 1992.

\bibitem[\protect\citeauthoryear{Serre \bgroup \em et al.\egroup
  }{2007}]{serre:2005}
T.~Serre, L.~Wolf, and T.~Poggio.
\newblock Object recognition with features inspired by visual cortex.
\newblock In {\em Proc. of Computer Vision and Pattern Recognition Conference},
  2007.

\bibitem[\protect\citeauthoryear{Simard \bgroup \em et al.\egroup
  }{2003}]{simard:2003}
P.~Simard, D.~Steinkraus, and J.~Platt.
\newblock Best practices for convolutional neural networks applied to visual
  document analysis.
\newblock In {\em Seventh International Conference on Document Analysis and
  Recognition}, pages 958--963, 2003.

\bibitem[\protect\citeauthoryear{Uetz and Behnke}{2009}]{uetz:2009}
R.~Uetz and S.~Behnke.
\newblock Large-scale object recognition with cuda-accelerated hierarchical
  neural networks.
\newblock In {\em IEEE International Converence on Intelligent Computing and
  Intelligent Systems (ICIS)}, 2009.

\bibitem[\protect\citeauthoryear{Wiesel and Hubel}{1959}]{wiesel:1959}
D.~H. Wiesel and T.~N. Hubel.
\newblock Receptive fields of single neurones in the cat's striate cortex.
\newblock {\em J. Physiol.}, 148:574--591, 1959.

\bibitem[\protect\citeauthoryear{Yu and Zhang}{2010}]{yu:2010}
K.~Yu and T.~Zhang.
\newblock Improved local coordinate coding using local tangents.
\newblock In {\em Proceedings of the International Conference on Machine
  Learning}, 2010.

\end{thebibliography}
\end{document}